\documentclass[a4paper,fleqn]{cas-dc}

\usepackage[numbers]{natbib}
\usepackage{microtype}
\usepackage{graphicx}
\usepackage{algorithm}
\usepackage{algpseudocode}
\usepackage{xcolor}
\usepackage{color}
\usepackage{tcolorbox}
\usepackage{CJK}
\usepackage{adjustbox}
\usepackage{float}
\usepackage{amsmath}
\usepackage{multirow}
\usepackage{subfigure}
\begin{document}
\let\WriteBookmarks\relax
\def\floatpagepagefraction{1}
\def\textpagefraction{.001}

\shorttitle{AED: An Black-box NLP Classifier Model Attacker}

\shortauthors{Yueyang Liu et~al.}

\title [mode = title]{AED: An Black-box NLP Classifier Model Attacker}                      
%\tnotemark[1,2]

%\tnotetext[1]{This document is the results of the research
%   project funded by the National Science Foundation.}

%\tnotetext[2]{The second title footnote which is a longer text matter
%   to fill through the whole text width and overflow into
%   another line in the footnotes area of the first page.}

\author[1]{Yueyang Liu}%[orcid=0000-0001-5894-8740]
[orcid=0000-0000-0000-0000]
\cormark[1]


%\ead{yliu114@student.gsu.edu}

%\credit{Conceptualization of this study, Methodology, Software}

\affiliation[1]{organization={Georgia State University},
    city={Atlanta},
    % citysep={}, % Uncomment if no comma needed between city and postcode
    postcode={30303}, 
    state={GA},
    country={USA}}

\author[2]{Yan Huang}

\author[1]{Zhipeng Cai}

\affiliation[2]{organization={Kennesaw State University},
    city={Kennesaw},
    % citysep={}, % Uncomment if no comma needed between city and postcode
    postcode={30144}, 
    state={GA},
    country={USA}}

\cortext[cor1]{Corresponding author}

%fntext[fn1]{This is the first author footnote. but is common to third
%  author as well.}
%\fntext[fn2]{Another author footnote, this is a very long footnote and
%  it should be a really long footnote. But this footnote is not yet
%  sufficiently long enough to make two lines of footnote text.}

%\nonumnote{This note has no numbers. In this work we demonstrate $a_b$
%  the formation Y\_1 of a new type of polariton on the interface
%  between a cuprous oxide slab and a polystyrene micro-sphere placed
%  on the slab.
%  }

\begin{abstract}
Deep Neural Networks (DNNs) have been successful in solving real-world tasks in domains such as connected and automated vehicles, disease , and job hiring. However, their implications are far-reaching in critical application areas. Hence, there is a growing concern regarding the potential bias and robustness of these DNN models. 
A transparency and robust model is always demanded in high-stakes domains where reliability and safety are enforced, such as healthcare and finance.
While most studies have focused on adversarial image attack scenarios, fewer studies have investigated the robustness of DNN models in natural language processing (NLP) due to their adversarial samples are difficult to generate.
To address this gap, we propose a word-level NLP classifier attack model called "AED," which stands for \textbf{\underline{A}}ttention mechanism enabled post-model \textbf{\underline{E}}xplanation with \textbf{\underline{D}}ensity peaks clustering algorithm for synonyms search and substitution.
AED aims to test the robustness of NLP DNN models by interpretability their weaknesses and exploring alternative ways to optimize them.
By identifying vulnerabilities and providing explanations, AED can help improve the reliability and safety of DNN models in critical application areas such as healthcare and automated transportation.
Our experiment results demonstrate that compared with other existing models, AED can effectively generate adversarial examples that can fool the victim model while maintaining the original meaning of the input. 
\end{abstract}

% \includegraphics{figs/grabs.pdf}

% \begin{highlights}
% \item We present a novel interpretable natural language classifier attack model based on a word score attention mechanism. It facilitates the interpretability ability of the victim model by producing detailed explanations in the form of text for every decision made by the models, also a minimum word perturbing number. 
% \item Using a density peaks clustering algorithm for word substitution: with a hierarchical search strategy proved synonyms candidates make the classifier misjudge while keeping humans unaware reduce the search area.
% \item We test our methodology using the IMDB and Amazon Review dataset; results indicate that our model is adequate to interpret the victim model; a minimum word substitution ratio can lead up to 20\% of miss judging to the victim.
% \end{highlights}

% Keywords

\maketitle

\section{Introduction}

Nowadays, Deep Neural Network (DNN) has become a prevalent domain and has been integrated into a wide range of applications because of the marvelous achievement in most fields, such as computer vision~\cite{esteva2021deep}, medical document understanding~\cite{nawroth2020emerging}, and connected autonomous vehicles~\cite{chen2019f}.
%motivation
Along with the enormous achievement of DNN, it is still hard to promote DNN technology to some critical industries, and the main resistance is the high concern of potential bias and robustness in these models~\cite{carlini2017towards,Explainabledang2022,bommasani2022opportunities}. To be specific, the DNN model is often referred to as the back box that lacks opacity and transparency and cannot explain how a specific decision was made. Meanwhile, the variants of the input data may cause unwanted prediction errors. 
For example, in medical records semantic analysis task, a false negative result concluded by an NLP model may ignore the critical note from the doctor that put the patient's life under threat. 
If an explainable model is applied here, the doctor could understand the basis for the algorithm’s diagnosis and provide necessary corrections.
Back to the natural language processing domain, previous research~\cite{liang2017deep,huang2017adversarial,boucher2022bad} also indicated that applied with perturbation strategies, the neural network classifier is likely to make prediction errors based on those perturbed samples, and even some state of art models are also inevitable.

%challenge
In order to improve the model robustness, a small variation of the training dataset will be applicable.~\cite{9741895} However, compared with other types of data, textual adversarial sample generation is more challenging; Due to their discrete feature, limited perturbation methods can be applied to text, and even the slightest character level change will lead to a grammatical error. 
Typically, based on the perturbation strategies, the textual adversarial sample generation can be classified into three levels: character, word, and sentence level~\cite{liang2017deep}. 
Many researchers have already shared their contributions to the word perturbation method, especially in search space creation and adversarial examples selection part~\cite{ren_generating_2019,zang_word-level_2020,liu_efficient_2021}.
However, to successfully fool the victim model, either their search space creates rely on a costly greedy search method or their models have no contribution to the model interpretable.

%ffl
To respond to those defects, in this article, we introduce a black-box word-level natural language classifier attack model called: "ADE".
To be special, to optimize the searching progress, we designed a word selection model based on \underline{A}ttention mechanism; with a similarity training approach, the model can assign scores to words related to its contribution to the classifier in a high-quality manner. The words assigned with a high score will be considered perturbation candidates.
Following a synonyms search algorithm selection algorithm based on \underline{D}ensity peaks clustering, where synonyms of target words that caused the largest prediction error will be chosen to generate an adversarial example. This strategy can minimize grammatical errors while rollover the prediction result.
In addition, via the generated attention score, the attack model can enable the victim model's \underline{E}xplainable feature by unveiling the decision-making theory, which can help the researchers correct their black box model's flaws and build user confidence and trust.

The main contribution of this work is listed as follows:
\begin{itemize}
\item We present a novel interpretable natural language classifier attack model based on a word score attention mechanism. It facilitates the interpretability ability of the victim model by producing detailed explanations in the form of text for every decision made by the models, also a minimum word perturbing number. 
\item Using a density peaks clustering algorithm for word substitution: with a hierarchical search strategy proved synonyms candidates make the classifier misjudge while keeping humans unaware reduce the search area.
\item We test our methodology using the IMDB and Amazon Review dataset; results indicate that our model is adequate to interpret the victim model; a minimum word substitution ratio can lead up to 20\% of miss judging to the target.
\end{itemize}

This paper is organized as follows. In section II, we explicate the backgrounds of major attention methods, word levels of NLP adversarial attack mechanisms, and model interpretability with a hybrid interpretable model approach. Section III introduces our attack model based on a word score attention mechanism and the density peaks clustering algorithm for word substitution. Section IV describes the attack experiment on two victim models with our methodology. Finally, Section V presents our conclusions and future work.

\section{Related works}

\subsection{Attention Mechanism}
In neural networks, attention is a technique that mimics cognitive attention.
The attention method was devised based on the human ability to quickly filter out valuable information with limited resources~\cite{ungerleider_mechanisms_2000}. For instance, instead of capturing all the senses, humans are likely to pay attention to some eye-catching parts only; by this means, with limited brain processing resources circumstances, humans still able to quickly select high-value information from massive noise.
This strategy significantly improves the human being perceptual external information procession's efficiency and accuracy.

Taking the advantage mentioned above, we see many researchers using the attention mechanism as an efficient resource allocation algorithm in their work.
Mnih et al.~\cite{mnih_recurrent_nodate} proposed an algorithm with an attention mechanism for image classification in 2014. 
While in their method, the model selects the input bounding box based on the past information and task demands rather than inputting the entire image.
While it can reduce time complexity, their result indicates that it can achieve remarkable results in image classification tasks. 
Bahdanau et al.~\cite{bahdanau_neural_2016} utilized the attention mechanism to the Neural Network Machine Translation (NMT) domain for the initial attempt. 
Their model can find the best position where the most information is concentrated. Then, the model predicts the target word based on the context vector associated with these source positions and all the previously generated target words. 
Fukui et al.~\cite{Fukui_2019_CVPR} proposed an Attention Branch Network (ABN), which extends a response-based visual explanation model by introducing a branch structure with an attention mechanism. Their research indicates that ABN outperforms baseline models on image recognition tasks, and an attention map is given for visual explanation.
Specifically, in the natural language processing domain, the attention mechanism has become an effective method to capture more informative information in the sequential corpus, while the text can be considered a product of mutual interaction between all sentences.~\cite{8908707}

In the Transformer model, Vaswani et al.~\cite{vaswani_attention_nodate} implemented a stacked self-attention and point-wise fully connected layers for both encoder and decoder cells rather than LSTM or CNN architectures.
In the translation tasks, the transformer can be trained significantly faster than architectures based on recurrent or convolutional layers.
A number of previous works have confirmed the effectiveness of the attention mechanism in sentiment analysis tasks~\cite{martins_softmax_nodate}. 
The attention mechanism can assign weights to features so that the classifier can use feature information in a focused manner.

More previous attempts~\cite{tan2017deep}~\cite{cao-etal-2018-adversarial}~\cite{yang2016hierarchical} and~\cite{vaswani_attention_nodate} render self-attention able to learn the dependencies between words in sentences and capture the inner structure information within sentences seen.
In this paper, we implement the Self-Interaction Attention Mechanism to our model; self-attention, also called intra Attention, is an attention mechanism relating different positions of a single sequence in order to compute a representation of the same sequence. It has been shown to be very useful in machine reading, abstractive summarization, or image description generation.
The model based on intra-attention gains a better perplexity score than LSTM or RNN model, and it has empirically shown to have an outstanding performance in sentiment analysis and Natural Language Inference (NLI) tasks~\cite{cheng_long_2016}. 
The self-attention mechanism's Query, Key, and Value originated from the identical input, which can be used to compute the attention distributions among the source, searching the dependency between each lexical item~\cite{lin_structured_2017} as:
\begin{equation}
\label{eq:attention}
Attention(Q, K, V) = \text{softmax}\left(\frac{QK^T}{\sqrt{d_k}}\right)V
\end{equation}
The attention mechanism computes a weighted sum of values, where the weights are determined by the softmax function applied to the dot product of the query and key matrices, scaled by the square root of the key dimension.
Q, K, and V are matrices representing queries, keys, and values, respectively
The scalar value $d_k$  is the dimensionality of the key and query matrices.
\subsection{Word-level Textual Adversarial Attack}
Commonly in the NLP model attack, we mainly applied perturbation strategies like insertion, modiﬁcation, and removal of sentences, words, or characters to fool the model to make the wrong production of those methods we call textual adversarial attacks.
According to the perturbations units, textual adversarial attacks can be divided into three levels according to the generated adversarial examples: character-level attacks, word-level attacks, and sentence-level. 
A character-level attack is to disturb several characters in a word, which can be the deletion, insertion, or swapping of two characters. The method work against currently-deployed commercial systems, however, misspellings can be easily detected during input sanitization~\cite{heigold_how_2017,boucher2022bad}. 

A word-level attack is to manipulate the whole word in the text. The most common word-level attacks are trigger phrase insertion~\cite{wallace2021concealed} and swapping~\cite{morris_textattack_2020}.
The modifications are mainly based on synonym substitution. Hence, after a careful design, humans are more imperceptible to those word-level modifications than char-level attacks. 

Meanwhile, according to the way of selecting manipulated words, the word-level adversarial attack can be classified into gradient-based, importance-based~\cite{gan_improving_2019}. 
Samanta and Mehta~\cite{samanta_towards_2017} proposed a word-level black-box attack based on word salience. 
This method first uses FGSM to approximate a word's contribution to the classification result. 
Then the adverb words that have a great contribution to the classifier will be substituted by a candidate. 
The candidate pool consists of synonyms, typos, and genre-specific keywords with high term frequencies in the corpus. Ren et al.~\cite{ren_generating_2019} improved Samanta and Mehta's method and gave out a greedy algorithm called probability-weighted word saliency (PWWS). 
The synonym set is built with WordNet, which means a candidate is selected if the substitute causes the most significant change in the classification probability. 
Alzantot et al.~\cite{alzantot_generating_2018} proposed a black box attack algorithm based on a genetic algorithm. In the algorithm, they provide a subroutine called Perturb. With a candidate pool with $k$ words to n maximize the target label prediction probability.

\subsection{Explainable AI}
In previous research, we see mainly NLP attack method researches are only focused on input corpus perturbing; however, it is crucial to understand which part contributes mostly to the model prediction fully. 
In this case, users can mathematically analyze the model input and output connection.
In our scenario, the victim model is a black-box frozen model, which did not satisfy any standards to consider an interpretable model. 

In order to enable its explainability of generating predictions for any provided input, some approaches referred to as post-modeling explainability are proposed~\cite{minh2021explainable}; some popular solution is feature relevance,
as an indirect approach to perform the post-modeling explainability, which evaluates the algorithm's internal processes by calculating the relevance score for all the variables that it manages. 
The computed score quantifies the importance or the sensitivity, which reveals what features are crucial, which is what the model depends on when making its prediction~\cite{Moradi_2021}.

In our method, a hybrid interpretable model approach~\cite{gallego2018clustering} is set to combine a black-box model with the victim interpretable model.
The basic theory is based on similarity learning but is different from common similarity learning to learning a similarity measure. In our trials, our attack model is learned to make a prediction the same as the frozen victim model, while a self-attention score layer generates a heat map to provide an explainability~\cite{payer2019integrating}.
This approach is based on the assumption: while we have two models, model $a$ is our attack model, and model $v$ is our pre-trained victim model. Given a corpus with features $w\in\{w_1, w_2\cdots w_n\}$ with a total $k$ category. 
In the pre-trained victim model, the model $f_v(w)$ map the data set to $k\in\{k_1, k_2\cdots k_i\}$ category with a probability of $P(f_v(w)=k_i)$, in the attack model $a$, the probability of $P(f_a(w)=k_i)$ is trained find the optimal $f_a$ with parameters $\theta$ that:
\begin{equation}
\label{eq:fa}
f_{a_{\theta}} = \underset{\theta}{\arg\max}\{P(f_a(w)=k_i) \sim P(f_v(w)=k_i)\}
\end{equation}
After optimization, the same corpus features input is expected to have the same predicted result, and we can 
replicate the black-box model $v$'s predictions with the replicated model $a$.

\section{Methodology}
After comparing different level NLP attacker's linguistic granularity, we introduce AED, an NLP black-box classifier attack model constituting of an attention score-based word-level rank algorithm, a density peaks clustering synonyms search algorithm. 
%Figure \ref{fig:The word ranking generator} illustrates the overall structure of our trainable model.

\subsection{Attention score-based word rank algorithm}
In the black box scenario, internal knowledge is not accessible.
To make our research feasible, we design an attack strategy consisting of word rank model training, which uses the observation of chosen inputs and predicts labels pair of the victim model to train the word rank model. 
When enough input and predicted pair exist, our model guarantees to compute a symmetric measure to the victim model.
This training method also can be considered as similarity learning. 
During the similarity learning training, we have the word rank model parameter $\theta$, data set $x$, and the prediction can be regarded as i.i.d. and follow a Bernoulli distribution.
The problem involves finding $\theta $ that best explains the data $x$: $P(x;\theta)$.
So we have:
\begin{equation}
\label{eq:mle}
\begin{aligned}
\hat{\theta}_{MLE} &= \underset{\theta}{\arg\max} \mathrm{P}(x \mid \theta) \\
&= \underset{\theta}{\arg\max} \prod_{i=1}^n \mathrm{P}(x_i \mid \theta) \\
&= \underset{\theta}{\arg\max} \sum_{i=1}^n \log \mathrm{P}(y_i \mid x_i, \theta)
\end{aligned}
\end{equation}

After training, the word rank model's parameters are expected to produce a max likely estimate of the victim model result.

\begin{figure}[h]
%left bottom right top
\centering
    \includegraphics[width=\linewidth,trim=9cm 2cm 9cm 1cm,clip]{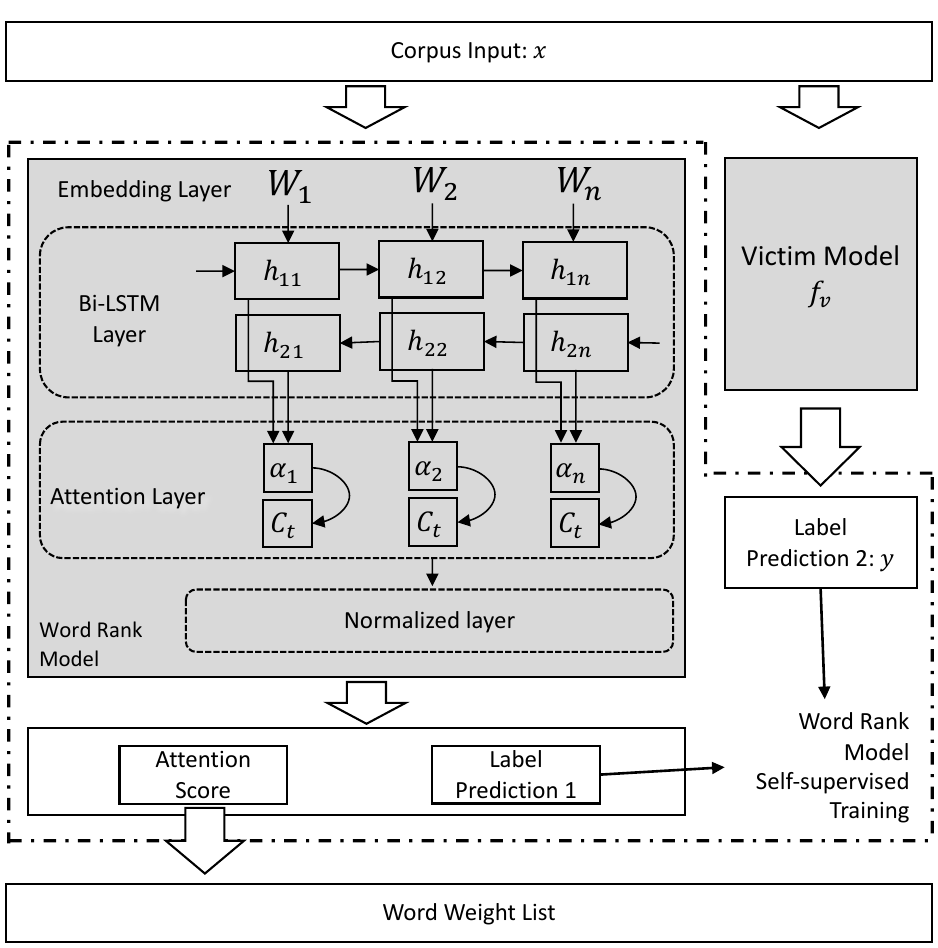}
\caption{Attention Score-based Word Rank Model - Word Rank Model replicates Victim Model's predictions and adds explainability to the replicated model.}
\label{fig:The word ranking generator}
\centering
\end{figure}

We elaborated our model structure in Figure \ref{fig:The word ranking generator}. Specifically, rather than specify a substituting location, we self-train our word rank model to compute the best substituting locations. Initially, retriever the corpus'$x$ related predict label $y$ via the victim model $y_i=f_v(x_i)$. Later, this prediction will be used as labels in word rank model self-training. This similarity learning strategy desires to enforce the relevance of the word rank and victim models.

To the structure of the score-based word rank model, it contains a word sequence encoder and a word-level self-attention layer. We use a two-layer Bidirectional LSTM~\cite{graves_framewise_2005} to obtain the hidden layer for the word sequence encoder part. 
After that, the self-attention layer provides a set of summation weight vectors for the LSTM hidden states. Weight vectors are dotted with the LSTM hidden states, which also work as an explanation method that aims to provide an understanding of why the system has arrived at this decision.
In our case, each corpus has a number of $n$ words, where each is defined as $x_t$, ($t \in n$) and embedded with the "GloVE.6B.100d" vector.
In the forward LSTM, we get the forward hidden state $h_{t1}=f_{\text{lstm}}(w_t,h_{t1-1}).$
To the backward LSTM with the same step, the backward hidden state can be represented as $h_{t2}=f_{\text{lstm}}(w_t,h_{t2-1}).$
Finally, concatenating the two-direction hidden state, the hidden output for the word vector $x_t$ will be $h_{t}=[h_{t1},h_{t2}]$. Meanwhile, define a context vector$s_t$ of Bi-LSTM.
Given a variable-length vector $a_i$ as weights, $c_t$is the context vector, and the self-attention is calculated as below:
\begin{equation}
\label{eq:ttt}
\begin{aligned}
c_t &= \sum_{i=1}^n \alpha_{i} h_i \\
\alpha_{i} &= \frac{\exp(\operatorname{score}(s_i,h_i))}{\sum_{i^{\prime}=1}^n \exp (\operatorname{score}(s_i, h_{i^{\prime}}))} \\
\operatorname{score}({s}_t, {h}_i) & = \mathbf{v^\top} \tanh (\mathbf{W}[{s}_t ,{h}_i])
\end{aligned}
\end{equation}
Where $\mathbf{v}$ is the Bi-LSTM output and $\mathbf{W}$ is the weight matrices to be learned in the model.
\subsection{The density peaks clustering synonyms search algorithm}
The density peaks clustering algorithm~\cite{rodriguez2014clustering} is based on the idea that cluster centers usually have a high density, and points surround them with lower density. Distance points with higher density usually have a large distance. 
Based on this criterion, first, we adopt the largest value and its corresponding word $\mathbf{word_i}$ in the weight matrices $\mathbf{W}$, which is given from the previous step; Meanwhile, according to the word embedding vectors, generate a word candidate pool with $\mathbf{word_i}$'s $K$ nearest neighbor  $\boldsymbol{X}=\left[\boldsymbol{x}_1, \boldsymbol{x}_2, \ldots, \boldsymbol{x}_K\right]^T$ in $M=100$ dimensional space.
One of the candidate words that cause the classifier the largest error will be chosen from the pool to substitute the word $\mathbf{word_i}$ in the original sentence. 
To guarantee the word similarity and provide a minimal disturbance, the distance matrix is computed via the word embedding method; for specific, GloVe vectors' Euclidean distance is denoted between the word $\mathbf{x_i}$ and the word $\mathbf{x_j}$, as follows:
\begin{equation}
\label{eq:distance}
\mathrm{d}\left(\boldsymbol{x}_i, \boldsymbol{x}_j\right)=\left\|\boldsymbol{x}_i-\boldsymbol{x}_{j_2}\right\|
\end{equation}
The local density $\rho$ of a word $\mathbf{x_i}$, is defined as:
\begin{equation}
\label{eq:rho_chi}
\begin{aligned}
\rho_i &=\sum_j \chi\left(\mathrm{d}\left(\boldsymbol{x}_i, \boldsymbol{x}_j\right)-d_c\right) \\
\chi(x) &= \begin{cases}1, & x<0 \\
0, & x \geq 0\end{cases}
\end{aligned}
\end{equation}
Regarding to the cutoff distance $d_c$, we choose the top 70\% as our threshold, which indicates that the distance distributed at the 30\% tail will be dropped.
The global distance $\delta$ between $\mathbf{x_i}$ and other higher density words is calculated as:
\begin{equation}
\label{eq:delta}
\delta_i=\left\{\begin{array}{l}
\min _{j: \rho_i>\rho_j}\left(\mathrm{~d}\left(\boldsymbol{x}_i, \boldsymbol{x}_j\right)\right), \text { if } \exists j \text { s.t. } \rho_i>\rho_j \\
\max _j\left(\mathrm{~d}\left(\boldsymbol{x}_i, \boldsymbol{x}_j\right)\right), \text { otherwise }
\end{array}\right.
\end{equation}

The candidate words with relatively high $\rho$ and high $\delta$ are considered cluster centers. After that remaining words to the same cluster as their nearest neighbors with higher density.
A hierarchical word list $HE$ will be created from each cluster, and the word within each cluster with the highest density and not in the hierarchical word list will be chosen. In our set, the hierarchical word list is limited to three layers; in distribution with $n$ cluster, the hierarchical word list will have $3*n$ words, and each layer has $n$ candidate.
In the searching step, iterated searching through the word list by layer, substituting the original word $\mathbf{word_i}$ with its candidate $w_i'$ to create an intermediate adversarial example $s'$. Then fed the sample into the victim classifier $f$; Here, we use a Cross-Entropy Loss as our loss function:

\begin{equation}
\label{eq:max_w}
\begin{aligned}
\arg \max_{w'_i \in HE} & -(f(\mathbf{word_i}) \cdot \log(f(w'_i)) \\
& +(1-f(\mathbf{word_i})) \cdot \log(1-f(w'_i)))
\end{aligned}
\end{equation}
Once the iterator finishes going through the hierarchical word list $HE$, or the victim model predicted label is reversed, the optimization step will regard as complete; the highest candidate word in the hierarchical word list that caused the highest loss value will be kept. 
%The over all model structure can be seen in the Algorithm~\ref{alg:alg1}. 
\begin{algorithm}
\renewcommand{\algorithmicrequire}{\textbf{Input:}}
\renewcommand{\algorithmicensure}{\textbf{Output:}}
\caption{Hierarchical word list Traversal}\label{alg:alg1}

\begin{algorithmic}
\Require original sample $s$, hierarchical word list $he$, victim and attack classifier $f_v$, $f_a$
\State $weight \gets f_a(s)$
\State $\sigma_0 \gets Loss(f_v(s))$
\State $wl \gets$ select words with top $k$ weight
\For{word $w$ in $wl$}
\State hierarchical word list $he \gets$ density peaks clustering
\For{each column $x$ in $he$}
\State candidates $w' \gets$ of $he[x][:]$
\State substituted sample $s' \gets$  $s[w]= w'_i$
\If{$\sigma(f_v(s'))>\sigma_0$} 
\State $\sigma_0 \gets \sigma(f_v(s'))$
\State $w \gets w'_i$ \Comment{substitute with the candidate}
\State $\upsilon_w \gets -\upsilon_w$ 
\While{$f_v(s) \neq f_v(s')$} 
\State\Return{$s'$} 
\Comment{stop iterate once success fool the victim}
\EndWhile
\EndIf
\EndFor
\EndFor
\State \Return{$s$} \Comment{failed fool the victim, return original}
\end{algorithmic}  
\end{algorithm}

It is important to notice that, compared with greedy search methods, our method uses attention weights and density peaks clustering to generate a hierarchical word list to limit the search area. The nearest neighbor of word stem embedding will guarantee the generated results' semantic similarity, but grammatical correctness is not assured here.
\section{Experiments}
In this section, we test our attack model on the IMDB and Amazon review dataset datasets. 
The following part proved the test setting and two victim model settings. 
We evaluate our model's performance from the attack success rate and measure the adversarial sentence similarity from the original.

\subsection{Experiments Set}
\noindent\textbf{Data Set}: We trained our model and the victim models using the IMDB dataset of movie reviews~\cite{maas_learning_nodate}. The IMDB dataset consists of 25,000 training examples and 25,000 test examples. Amazon review dataset~\cite{DBLP:journals/corr/ZhangZL15} contains 3,600,000 phrases with fine-grained sentiment labels in the parse trees of 400,000 sentences from product reviews and metadata from Amazon. Due to the limitation, we randomly select 30,000 phrases and 20,000 samples as our sub-date set.

\noindent\textbf{Victim Models}: We train a CNN model, a GRU model, and a Bert model as our victim model. 
We limit the word dictionary length to 30000. 
The CNN model consists of an embedding layer that performs 100-dimensional word embedding on 2500-dimensional input vectors, two 1D-convolutional layers, a 1D-max-pooling layer, and two fully connected layers. 
The GRU model includes a 2-layer of recurrent layer and embedding dimensions set to 100. 
In both models, we use Adam as our optimizer. 

Initially, we train the victim model on both of the data sets,
the GRU achieved an 89.63\% accuracy rate in unmodified IMDB data and 95.96\% accuracy in the unmodified Amazon data set, while the CNN model has an accuracy of 86.97\% and 93.54\%, respectively. Noteworthy, we also applied a BERT model here; however, due to the resource limit, the BERT model resulted in no significant accuracy improvement after a long time of training, so we did not involve the BERT model in the next step.
After the victim model is fully trained, our attack model will query the victim model continuously and be self-trained to make similar perdition and an attention weight list for each corpus. 
\subsection{Evaluation}
In this section, we evaluate our method on the test dataset. The attack model is expected to trick the victim model into making a wrong prediction by the generation of adversarial input.
In each epoch, depending on the weight list provided by the attack model, the word which contributes most to the classifier will be selected and replaced by the one candidate from the hierarchical synonyms word list.

\begin{figure}[h]
\centering
\includegraphics[width=\linewidth]{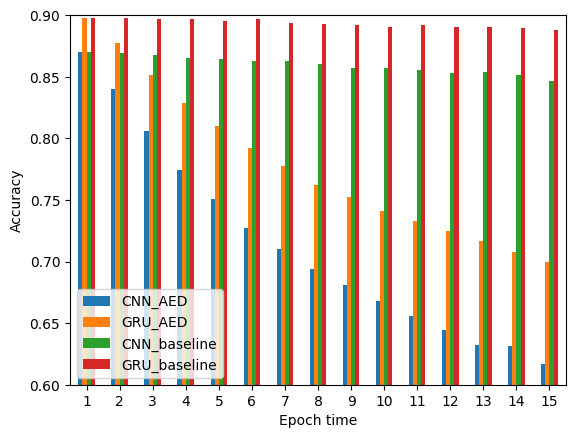}
\caption{The accuracy of the IMDB dataset. The figure shows the accuracy of the two victim models (CNN/GRU) on the IMDB dataset with the AED attack model and random word replace attack(baseline).}
\label{fig:IMDB ATTACK}
\end{figure}

\begin{figure}[h]
\centering
\includegraphics[width=\linewidth]{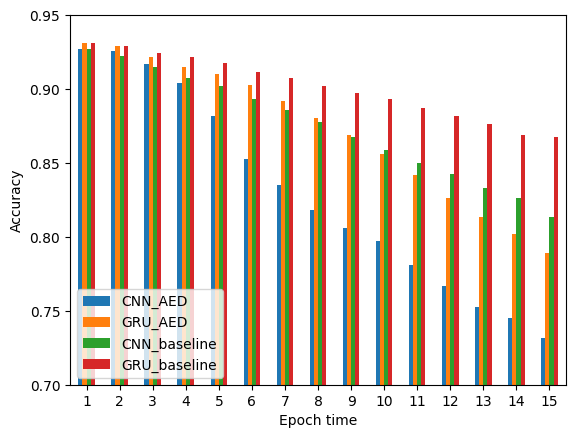}
\caption{The accuracy of the Amazon review dataset. The figure shows the accuracy of the two victim models (CNN/GRU) on the Amazon review dataset with the AED attack model and random word replace attack(baseline).}
\label{fig:AZ ATTACK}
\end{figure}

\noindent\textbf{Success Rate}: We totally conduct 15 epochs in our paper, and in each epoch, we measure the accuracy of the victim model on the manipulated input data. The success rate of the attack algorithm is defined as the percentage of the wrong prediction by the two victim models. Higher success rates indicate that the attack model can generate more destructive adversaries samples that can cause the victim model inversion results.
The experiment result shows in Figure\ref{fig:IMDB ATTACK}, and Figure\ref{fig:AZ ATTACK} show that within 15 epochs, our attack model caused up to 20\% of a downgrade of the victim model accuracy; compared with our baseline model: word substitution with random select strategy, at the same epoch, the attack model has a higher percentage of success rate.

Since the more effective the attacking method is, the more the classification accuracy of the model drops. Table~\ref{tab:Cases & Accuracy} illustrates the classiﬁcation accuracy of different models on the original samples and the adversarial samples generated by the different numbers of words substituted. 

\begin{table*}[h]
%\tablefont%
\begin{center}\caption{Test Attack Success Rate of victim model}\label{tab:Cases & Accuracy}
\begin{tabular}{c|c|cc|c|c|ccc}
\hline\hline
Substitution Method & \multicolumn{4}{c|}{Random} & \multicolumn{4}{c}{AED: Attention \& density peaks} \\ \hline
Victim model & GRU & \multicolumn{1}{c|}{CNN} & GRU & CNN & GRU & \multicolumn{1}{c|}{CNN} & \multicolumn{1}{c|}{GRU} & CNN \\ \hline
Epoch & \multicolumn{2}{c|}{IMDB} & \multicolumn{2}{c|}{Amazon review} & \multicolumn{2}{c|}{IMDB} & \multicolumn{2}{c}{Amazon review} \\ \hline
1 & 89.63 & \multicolumn{1}{c|}{86.97} & 95.96 & 93.54 & 89.77 & \multicolumn{1}{c|}{87.04} & \multicolumn{1}{c|}{93.14} & 92.71 \\
5 & 89.53 & \multicolumn{1}{c|}{86.47} & 91.75 & 90.19 & 81.02 & \multicolumn{1}{c|}{75.06} & \multicolumn{1}{c|}{91.00} & 88.15 \\
10 & 89.06 & \multicolumn{1}{c|}{85.70} & 89.34 & 85.90 & 74.08 & \multicolumn{1}{c|}{66.79} & \multicolumn{1}{c|}{85.62} & 79.73 \\
15 & 88.81 & \multicolumn{1}{c|}{84.68} & 86.75 & 81.33 & 69.93 & \multicolumn{1}{c|}{61.69} & \multicolumn{1}{c|}{78.94} & 73.18 \\ \hline\hline
\end{tabular}
\end{center}
\end{table*}

In addition to evaluating the effectiveness of our approach in generating adversarial samples, we also explored methods for enhancing the explainability of victim models. 
To this end, we utilized a heat map technique based on the weight list presented in Table\ref{tab:examples}, in which the contribution of each word to the final prediction was indicated by color.
This allowed us to observe the impact of each word on the final prediction, providing a valuable tool for interpreting the model's decision-making process and helping the human evaluation during the generated sample evaluation part.

\begin{table*}[!hb]
\caption{Adversarial examples generated from the ADE model - The output is a heat map showing each word's contribution, with deep red representing larger weight values, while the green color is the synonyms substitution.}
\label{tab:examples}
\begin{center}
\begin{tabular}{c|c|c}
\hline\hline
Text example & {Original} & \multicolumn{1}{c}{Adversarial} \\ \hline
\multirow{9}{*}{{\setlength{\fboxsep}{0pt}\colorbox{white!0}{\parbox{0.63\textwidth}{
\colorbox{red!0.0}{\strut If} \colorbox{red!0.00012049153}{\strut only} \colorbox{red!2.4076482}{\strut to} \colorbox{red!41.850613}{\strut avoid} \colorbox{red!2.6094491}{\strut making} \colorbox{red!2.762422}{\strut this} \colorbox{red!3.3708293}{\strut type} \colorbox{red!2.4282415}{\strut of} \colorbox{red!0.048248764}{\strut film} \colorbox{red!0.8618093}{\strut in} \colorbox{red!3.2251544}{\strut the} \colorbox{red!0.48469606}{\strut future} \colorbox{red!0.40476096}{\strut .} \colorbox{red!0.10049809}{\strut This} \colorbox{red!0.007725288}{\strut film} \colorbox{red!0.111630626}{\strut is} \colorbox{red!100}{\strut interesting} \colorbox{green!100}{\strut fascinating} \colorbox{red!11.977647}{\strut as} \colorbox{red!0.7110907}{\strut an} \colorbox{red!99.0835}{\strut experiment} \colorbox{green!99.0835}{\strut analysis} \colorbox{red!8.042701}{\strut but} \colorbox{red!0.36761093}{\strut tells} \colorbox{red!15.277857}{\strut no} \colorbox{red!5.5886183}{\strut cogent} \colorbox{red!0.70954347}{\strut story} \colorbox{red!0.48932195}{\strut .} \colorbox{red!0.93683857}{\strut One} \colorbox{red!49.501488}{\strut might} \colorbox{red!6.1803293}{\strut feel} \colorbox{red!82.96171}{\strut virtuous} \colorbox{green!82.96171}{\strut honest} \colorbox{red!0.120363995}{\strut for} \colorbox{red!10.417601}{\strut sitting} \colorbox{red!5.6367707}{\strut thru} \colorbox{red!0.55203253}{\strut it} \colorbox{red!36.305832}{\strut because} \colorbox{red!3.2666464}{\strut it} \colorbox{red!23.586298}{\strut touches} \colorbox{red!0.9998154}{\strut on} \colorbox{red!24.52306}{\strut so} \colorbox{red!2.0461967}{\strut many} \colorbox{red!5.8286934}{\strut IMPORTANT} \colorbox{red!0.23417546}{\strut issues} \colorbox{red!2.771838}{\strut but} \colorbox{red!1.9464558}{\strut it} \colorbox{red!4.7967668}{\strut does} \colorbox{red!89.804695}{\strut so} \colorbox{green!89.804695}{\strut extremely} \colorbox{red!12.544616}{\strut without} \colorbox{red!34.995117}{\strut any} \colorbox{red!5.5647287}{\strut discernable} \colorbox{red!36.45725}{\strut motive} \colorbox{red!0.08867267}{\strut .} \colorbox{red!0.00083411473}{\strut The} \colorbox{red!0.39388826}{\strut viewer} \colorbox{red!5.995576}{\strut comes} \colorbox{red!70.15161}{\strut away} \colorbox{red!1.6910005}{\strut with} \colorbox{red!36.302094}{\strut no} \colorbox{red!0.079994865}{\strut new} \colorbox{red!0.006144634}{\strut perspectives} \colorbox{red!0.013095121}{\strut (} \colorbox{red!24.020319}{\strut unless} \colorbox{red!1.0920794}{\strut one} \colorbox{red!4.6411586}{\strut comes} \colorbox{red!12.764788}{\strut up} \colorbox{red!1.7333813}{\strut with} \colorbox{red!2.9285157}{\strut one} \colorbox{red!4.3437123}{\strut while} \colorbox{red!1.729867}{\strut one} \colorbox{red!1.4261452}{\strut 's} \colorbox{red!2.092836}{\strut mind} \colorbox{red!99.44868}{\strut wanders} \colorbox{green!99.44868}{\strut amble} \colorbox{red!0.20848742}{\strut ,} \colorbox{red!0.5010061}{\strut as} \colorbox{red!2.2797732}{\strut it} \colorbox{red!0.10162068}{\strut will} \colorbox{red!68.414474}{\strut invariably} \colorbox{red!4.987791}{\strut do} \colorbox{red!1.8871096}{\strut during} \colorbox{red!0.74444944}{\strut this} \colorbox{red!100}{\strut pointless} \colorbox{green!100}{\strut ineffectual} \colorbox{red!0.12278526}{\strut film} \colorbox{red!0.011520135}{\strut )} \colorbox{red!0.40079227}{\strut .} \colorbox{red!2.7248492}{\strut One} \colorbox{red!39.51815}{\strut might} \colorbox{red!7.0996614}{\strut better} \colorbox{red!2.4185164}{\strut spend} \colorbox{red!0.36762214}{\strut one} \colorbox{red!0.6932021}{\strut 's} \colorbox{red!0.74395263}{\strut time} \colorbox{red!99.2929}{\strut staring} \colorbox{green!99.2929}{\strut open} \colorbox{red!3.1187801}{\strut out} \colorbox{red!1.256041}{\strut a} \colorbox{red!1.1464034}{\strut window} \colorbox{red!0.16495734}{\strut at} \colorbox{red!0.44934747}{\strut a} \colorbox{red!1.8475695}{\strut tree} \colorbox{red!3.2202246}{\strut growing}
}}}} & \multirow{5}{*}{Negative} & \multirow{5}{*}{Positive} \\
 &  &  \\
 &  &  \\
 &  &  \\
 &  &  \\
 & \multirow{5}{*}{0.139} & \multirow{5}{*}{0.723} \\
 &  &  \\
 &  &  \\
 &  &  \\ \hline
\multirow{12}{*}{{\setlength{\fboxsep}{0pt}\colorbox{white!0}{\parbox{0.63\textwidth}{
\colorbox{red!0.6565636}{\strut 
} \colorbox{red!0.013150538}{\strut Originally} \colorbox{red!0.006655227}{\strut I} \colorbox{red!0.5333303}{\strut was} \colorbox{red!5.196091}{\strut a} \colorbox{red!40.794956}{\strut Tenacious} \colorbox{red!0.15005109}{\strut D} \colorbox{red!0.23159727}{\strut fan} \colorbox{red!0.08679423}{\strut of} \colorbox{red!0.45531365}{\strut their} \colorbox{red!0.37195474}{\strut first} \colorbox{red!0.89807934}{\strut album} \colorbox{red!0.84212136}{\strut and} \colorbox{red!11.303427}{\strut naturally} \colorbox{red!2.264821}{\strut listened} \colorbox{red!0.09717623}{\strut to} \colorbox{red!0.10220015}{\strut a} \colorbox{red!3.7088368}{\strut few} \colorbox{red!0.017133728}{\strut tracks} \colorbox{red!0.0111853015}{\strut off} \colorbox{red!0.0}{\strut The} \colorbox{red!0.009029282}{\strut P.O.D.} \colorbox{red!1.7860081}{\strut and} \colorbox{red!6.240959}{\strut was} \colorbox{red!11.32491}{\strut rather} \colorbox{red!99.74}{\strut disappointed} \colorbox{green!99.74}{\strut discouraged} \colorbox{red!11.224195}{\strut .} \colorbox{red!0.35408965}{\strut After} \colorbox{red!3.4043915}{\strut watching} \colorbox{red!7.9360204}{\strut the} \colorbox{red!3.714546}{\strut movie} \colorbox{red!3.0262911}{\strut ,} \colorbox{red!48.814186}{\strut my} \colorbox{red!34.555504}{\strut view} \colorbox{red!4.4628153}{\strut was} \colorbox{red!19.499346}{\strut changed} \colorbox{red!0.3460458}{\strut .} \colorbox{red!0.003926771}{\strut The} \colorbox{red!0.21556549}{\strut movie} \colorbox{red!6.970563}{\strut is} \colorbox{red!99.154}{\strut pretty} \colorbox{green!99.154}{\strut very} \colorbox{red!99.6973}{\strut funny} \colorbox{green!99.6973}{\strut amusing} \colorbox{red!8.054149}{\strut from} \colorbox{red!1.5187606}{\strut beginning} \colorbox{red!0.18949993}{\strut to} \colorbox{red!0.12666748}{\strut the} \colorbox{red!0.49346402}{\strut end} \colorbox{red!0.2484026}{\strut and} \colorbox{red!0.40262607}{\strut found} \colorbox{red!1.5464401}{\strut my} \colorbox{red!0.49172595}{\strut self} \colorbox{red!0.14131123}{\strut engaged} \colorbox{red!0.20444255}{\strut in} \colorbox{red!0.32707646}{\strut it} \colorbox{red!14.475655}{\strut even} \colorbox{red!0.8257934}{\strut though} \colorbox{red!0.60261077}{\strut it} \colorbox{red!0.7452609}{\strut was} \colorbox{red!59.740242}{\strut really} \colorbox{red!15.877559}{\strut was} \colorbox{red!17.534952}{\strut a} \colorbox{red!99.2415}{\strut stupid} \colorbox{green!99.2415}{\strut dull}  \colorbox{red!57.8403}{\strut storyline} \colorbox{red!37.01881}{\strut because} \colorbox{red!3.7242377}{\strut of} \colorbox{red!6.9792504}{\strut the} \colorbox{red!9.156061}{\strut attitudes} \colorbox{red!1.9449528}{\strut that} \colorbox{red!3.1653636}{\strut KG} \colorbox{red!2.8952086}{\strut and} \colorbox{red!3.5867937}{\strut Jaybles} \colorbox{red!12.951551}{\strut portray} \colorbox{red!4.5742483}{\strut in} \colorbox{red!25.140139}{\strut the} \colorbox{red!28.088587}{\strut movie} \colorbox{red!3.5426354}{\strut .} \colorbox{red!1.1547828}{\strut Much} \colorbox{red!99.30299}{\strut more} \colorbox{green!99.30299}{\strut extra} \colorbox{red!99.012}{\strut entertaining} \colorbox{green!99.012}{\strut affecting} \colorbox{red!99.19682}{\strut and} \colorbox{red!82.155}{\strut enjoyable} \colorbox{green!82.155}{\strut affecting} \colorbox{red!90.9135}{\strut than} \colorbox{red!0.8735515}{\strut movies} \colorbox{red!0.012306343}{\strut I} \colorbox{red!0.53085756}{\strut have} \colorbox{red!11.724824}{\strut seen} \colorbox{red!2.1491363}{\strut in} \colorbox{red!1.6206634}{\strut the} \colorbox{red!0.95896524}{\strut theaters} \colorbox{red!20.701717}{\strut lately} \colorbox{red!0.13461111}{\strut .} \colorbox{red!0.003323231}{\strut ex} \colorbox{red!0.0024767711}{\strut .} \colorbox{red!0.00021468058}{\strut Saw} \colorbox{red!0.079413354}{\strut III} \colorbox{red!0.44150186}{\strut (} \colorbox{red!98.329}{\strut dull} \colorbox{red!6.6723313}{\strut and} \colorbox{red!26.744858}{\strut dragging} \colorbox{red!0.21512145}{\strut )} \colorbox{red!2.2129133}{\strut ,} \colorbox{red!1.5892271}{\strut Casino} \colorbox{red!97.01838}{\strut Royale} \colorbox{red!0.42113933}{\strut (} \colorbox{red!33.707737}{\strut way} \colorbox{red!1.9151225}{\strut to} \colorbox{red!0.3203026}{\strut homo} \colorbox{red!0.101104036}{\strut -} \colorbox{red!0.63212216}{\strut erotic} \colorbox{red!0.051467136}{\strut )} \colorbox{red!0.5746424}{\strut which} \colorbox{red!10.872552}{\strut in} \colorbox{red!9.08564}{\strut prior} \colorbox{red!0.3276908}{\strut installments} \colorbox{red!0.037045218}{\strut I} \colorbox{red!26.268368}{\strut have} \colorbox{red!99.0529}{\strut really} \colorbox{green!99.0529}{\strut certainly} \colorbox{red!100}{\strut enjoyed} \colorbox{green!100}{\strut like} \colorbox{red!14.173699}{\strut If} \colorbox{red!70.39257}{\strut you} \colorbox{red!84.41216}{\strut enjoyed} \colorbox{red!0.29850116}{\strut Borat} \colorbox{red!1.6781433}{\strut ,} \colorbox{red!22.124388}{\strut you} \colorbox{red!6.846855}{\strut will} \colorbox{red!92.50415}{\strut enjoy} \colorbox{red!7.685054}{\strut the} \colorbox{red!26.623219}{\strut tale} \colorbox{red!0.015938215}{\strut of} \colorbox{red!1.479662}{\strut The} \colorbox{red!0.0017198063}{\strut Greatest} \colorbox{red!1.6228628}{\strut Band} \colorbox{red!3.677963}{\strut on} \colorbox{red!0.4726726}{\strut Earth} \colorbox{red!0.18236093}{\strut 
}
}}}} & \multirow{6}{*}{Positive} & \multirow{6}{*}{Positive} \\
 &  &  \\
 &  &  \\
 &  &  \\
 &  &  \\
 & \multirow{6}{*}{0.852} & \multirow{6}{*}{0.379} \\
 &  &  \\
 &  &  \\
 &  &  \\ 
 &  &  \\
 &  &  \\ \hline\hline
\end{tabular}
\end{center}
\end{table*}

\noindent\textbf{Sentence Similarity}: To evaluate the adversarial example quality~\cite{farouk2019measuring}, we considered  three methods to evaluate the generated adversarial examples' quality. 

First,  calculate the \textit{Pearson Correlation Coefficient} between the initial sample and the generated adversarial examples after each iterator. After iterating time $k$, we have the original sample matrix $m_0$, and adversarial examples $m_n$; each matrix has a size $n$. To simplify, we have the equation below: 
\begin{equation}
\label{eq:rho_k}
\rho_k=\overline{\sum|1-\frac{6 \sum (m_0-m_n)^{2}}{n^{3}-n}|}.
\end{equation}

Second, we consider evaluating the \textit{Semantic Textual Similarity} with machine learning between the initial sample and the generated adversarial examples after each iterator.  At the time of writing this paper, we use the model "stsb-roberta-large"\cite{reimers2019sentencebert}; the final similarity score is present in cosine similarity.

\begin{figure}[h!]
\centering\includegraphics[width=\linewidth]{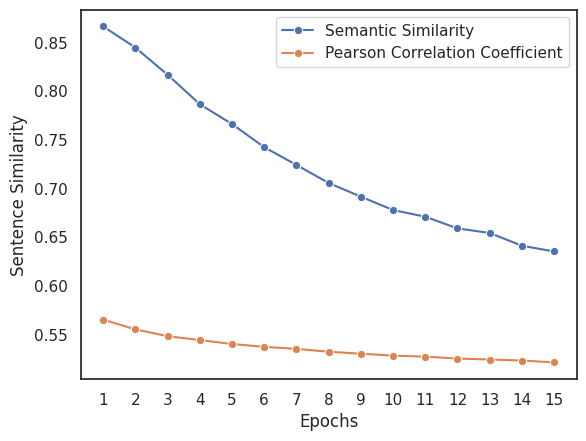}
\centering
\caption{Sentence Similarity. The similarity score is calculated on Pearson Correlation Coefficient \& Semantic Similarity}
\label{fig:Sentence Similarity}
\centering
\end{figure}

Figure~\ref{fig:Sentence Similarity} illustrates the semantic textual similarity and Pearson correlation coefficient after each iteration. After 15 epochs, the Semantic Textual Similarity score went downward from 0.99 to 0.65, while the minimum coefficient of generated examples after each iterator remains over 0.5. Considering the generated quality and performance, we believe our method can achieve the trade-off point after seven or eight words substitution.

Nevertheless, we drove out some samples used for \textit{Human Evaluation}.
In Table~\ref{tab:examples}, the heat map indicates the word attention score. 
The deeper the color, the higher value the word got. The two samples are the results after we substituted the top 7 words and 9 words and successfully made the victim model make the wrong prediction: 0.139 (negative) to 0.723 (positive) and 0.852 (positive) to 0.379 (negative) separately. 

Compared with Alzantot et al.~\cite{alzantot_generating_2018} Ren et al.~\cite{ren_generating_2019}, our method offers a notable advantage of being more cost-effective. Specifically, our technique involves density peaks clustering and hierarchical synonyms search only for the nearest neighbor of words that score higher, which significantly reduces the computational burden. The results of our experiments demonstrate that our method is highly effective in reducing text classification accuracy, even at a low substitution rate. Moreover, the incorporation of an attention mechanism in our approach enhances the model's explainability, allowing for greater interpretability of the generated adversarial samples.

\section{Conclusions and Future Work}

In this paper, we proposed an NLP classifier attack method that includes an attention mechanism-based word score method and a density peaks clustering algorithm for word substitution, which can generate imperceptible adversarial examples. 
In the corpus, the words are given an order by the word saliency and weighted by the classification probability; after that conduct word substitution from the density peaks cluster. 

Experiments show that our methods can greatly reduce text classification accuracy with a low word substitution rate, and such perturbation is hard for humans to perceive. 
Comparison with existing baselines shows the advantage of our method. Our model proposed in this paper can work with a variety type of NLP models to help improve their query efficiency and success rates. 

On the other hand, by using the attention score, our model can enhance the explainability of the black-box model; other researchers can use this to help find the vulnerability, unfairness, and safety issues in NLP models and therefore help convince the users to entrust applications that are based on those model in order to make crucial decisions.

Despite the promising results of our method, we acknowledge that there are limitations to our study. Specifically, we were only able to manually verify a few generated adversarial examples due to time constraints. Additionally, we opted to abandon the use of the Bert model in our experiments due to the trade-off between training time and computational resources.

Meanwhile, our evaluation of the semantic accuracy of synonyms revealed that some synonyms were not as accurate as their original counterparts. While our approach was successful in reducing text classification accuracy, the Pearson correlation coefficient and semantic textual similarity indicated that there were instances where the semantics of the generated synonyms did not fully align with the original words. As such, we recognize the need for more extensive and elaborative human evaluation to determine the overall effectiveness of our method.

Further research is needed to explore the full potential of our approach, including evaluating its performance in a broader range of natural language processing tasks and scenarios. Then, conduct a grammar check for the final result. Nonetheless, we remain confident that our method represents a significant step forward in adversarial attacks and offers valuable insights for improving the security and robustness of NLP models.

\printcredits

\bibliographystyle{cas-model2-names}

\bibliography{cas-refs}


\clearpage 

\end{document}